\documentclass[10pt,journal,compsoc]{IEEEtran}
\usepackage[noadjust]{cite}
\usepackage{float}
\usepackage{bm}
\usepackage{graphicx}
\usepackage{amsmath}
\usepackage{amsthm}
\usepackage{amsfonts}
\usepackage{algpseudocode}
\usepackage{algorithmicx}
\usepackage{rotating}
\usepackage{multirow}
\usepackage{url}
\usepackage[caption=false,font=normalsize,labelfont=sf,textfont=sf]{subfig}
\usepackage{threeparttable}
\usepackage{hyperref}

\usepackage{times}
\usepackage{amssymb}
\usepackage{color}
\usepackage{rotating}

\DeclareMathOperator*{\argmax}{arg\,max}
\DeclareMathOperator*{\argmin}{arg\,min}

\newtheorem{dfn}{Definition}

\begin{document}

\title{Boosting with Lexicographic Programming: Addressing Class Imbalance without\\ Cost Tuning}

\author{Shounak~Datta, Sayak~Nag, and Swagatam~Das%
\IEEEcompsocitemizethanks{\IEEEcompsocthanksitem Shounak Datta (Email: shounak.jaduniv@gmail.com) and Swagatam Das (Email: swagatam.das@isical.ac.in) are with the Electronics and Communication Sciences Unit, Indian Statistical Institute, Kolkata, India. \protect
\IEEEcompsocthanksitem Sayak~Nag was formerly with the Instrumentation \& Electronics Engineering Department, Jadavpur University, Salt Lake Campus, Kolkata, India. Email: sayak.nag9@gmail.com. \protect
\IEEEcompsocthanksitem Corresponding author: Swagatam Das
\IEEEcompsocthanksitem This paper has supplementary downloadable material for online publication only, as provided by the authors. This includes details of the experimental results summarized in this article and is xxx KB in size.}
}

\makeatletter
\long\def\@IEEEtitleabstractindextextbox#1{\parbox{0.922\textwidth}{#1}}
\makeatother

\IEEEtitleabstractindextext{
\begin{abstract}
A large amount of research effort has been dedicated to adapting boosting for imbalanced classification. However, boosting methods are yet to be satisfactorily immune to class imbalance, especially for multi-class problems. This is because most of the existing solutions for handling class imbalance rely on expensive cost set tuning for determining the proper level of compensation. We show that the assignment of weights to the component classifiers of a boosted ensemble can be thought of as a game of \emph{Tug of War} between the classes in the margin space. We then demonstrate how this insight can be used to attain a good compromise between the rare and abundant classes without having to resort to cost set tuning, which has long been the norm for imbalanced classification. The solution is based on a lexicographic linear programming framework which requires two stages. Initially, class-specific component weight combinations are found so as to minimize a hinge loss individually for each of the classes. Subsequently, the final component weights are assigned so that the maximum deviation from the class-specific minimum loss values (obtained in the previous stage) is minimized. Hence, the proposal is not only restricted to two-class situations, but is also readily applicable to multi-class problems. Additionally, we also derive the dual formulation corresponding to the proposed framework. Experiments conducted on artificial and real-world imbalanced datasets as well as on challenging applications such as hyperspectral image classification and ImageNet classification establish the efficacy of the proposal.
\end{abstract}

\begin{IEEEkeywords}
Boosting, Imbalanced classification, Lexicographic Linear Programming, Cost set tuning, Multi-Criterion Decision Making
\end{IEEEkeywords}
}

\maketitle
\IEEEpeerreviewmaketitle

\section{Introduction}\label{sec:intro}

\IEEEPARstart{B}{oosting} \cite{schapire1990strength} is an ensemble learning technique that operates by repeatedly training a so-called weak classifier on reweighted versions of the basic dataset. The reweighting is done so that data instances misclassified in the previous round are assigned greater weights in the current round. The variants thus trained become the component classifiers of the ensemble having weightage proportional to their performance on the training data. Boosting is known to exhibit resistance to overfitting for noise-free datasets, owing to its ability to optimize the margin of the underlying weighted combination of weak classifiers \cite{schapire1998boosting}.

However, boosting methods (and classification techniques in general) are unable to properly handle datasets characterized by class imbalance of data, i.e. when one or more (but not all) classes in the dataset have a small number of representatives in the training sample. Such imbalanced or uneven datasets often arise in critical real life applications such as medical diagnosis \cite{mazurowski2008training}, fraud detection \cite{phua2004minority}, etc. In fact, even the state-of-the-art deep-learning techniques which are used for complex computer vision applications, suffer due to class imbalance \cite{huang16cvpr}. Nikolaou et al. \cite{nikolaou2016cost}, in a rather exhaustive comparative study, observe that a significant amount of research effort has been aimed towards adapting boosting methods like AdaBoost \cite{freund1995desicion} for such class imbalanced learning tasks. Despite continued research efforts \cite{ting2000comparative, viola2002fast, sun2007cost, masnadi2007asymmetric, masnadi2011cost, galar2012, landesa2013double, wang2016online, ohsaki2017confusion}, classification methods in general and boosting methods in particular have yet to become sufficiently immune to class imbalance due to the following reasons:
\begin{itemize}
\item Most of the boosting variants proposed to handle class imbalance \cite{ting2000comparative, masnadi2007asymmetric, nikolaou2015calibrating, nikolaou2016cost} assume that the relative costs of misclassifying the two classes are known \emph{a priori}. This is often not true and the set of relative costs that are most suitable for a particular dataset must be found by a costly parameter tuning regime.
\item Moreover, most of the research efforts have been aimed at handling dichotomous or the so-called two-class imbalanced problems, with little attention being accorded to multi-class or polychotomous classification problems characterized by class imbalance. One of the reasons behind this is the need for parameter tuning, which becomes costlier for multi-class datasets. This is because, in a multi-class situation, either an entire matrix of costs must be estimated \cite{zhou2010multi}, or multiple relative costs (pertaining to multiple two-class problems obtained by decomposing the multi-class problem) must be tuned \cite{krawczyk2016cost}.
\item Moreover, imbalanced classification on multi-class datasets is further complicated by the fact that the class imbalance can be of multi-minority type (the case where multiple classes are underrepresented), multi-majority type (the case where only a single class has significantly lower number of representatives compared to other classes), or a combination thereof \cite{wang2012multiclass}.
\end{itemize}

\begin{figure*}[!t]
\centering
\subfloat[Toy class imbalanced dataset with outliers in both classes.]{\includegraphics[width=0.5\textwidth]{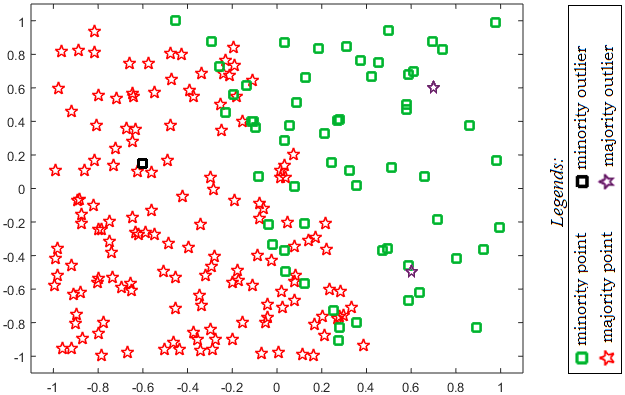}
\label{figMotivData}}
\hfill
\subfloat[Round 1.]{\includegraphics[width=0.27\textwidth]{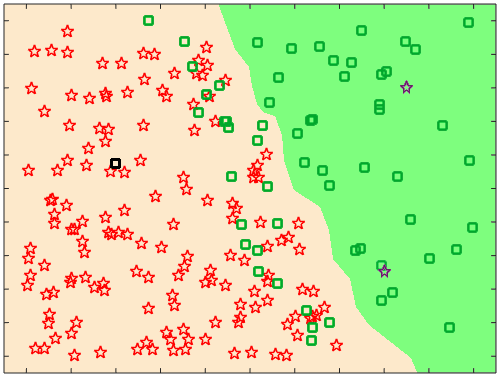}
\label{figMotiv1}}
\subfloat[Round 2.]{\includegraphics[width=0.27\textwidth]{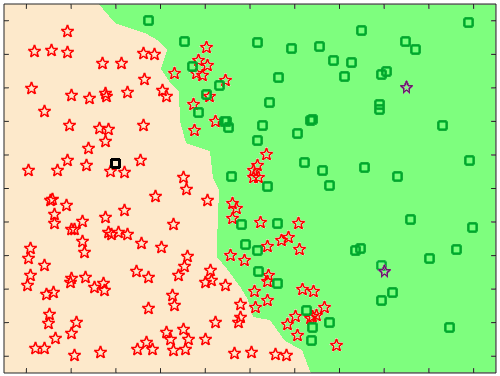}
\label{figMotiv2}}
\subfloat[Round 3.]{\includegraphics[width=0.27\textwidth]{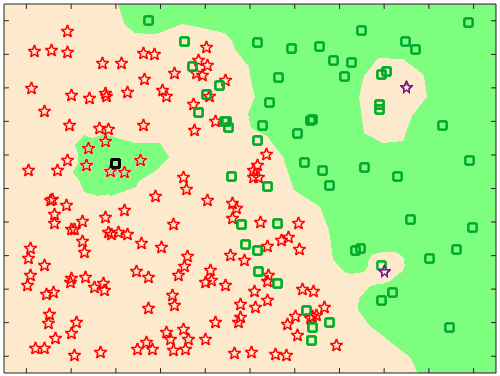}
\label{figMotiv3}}
\caption{Toy class imbalanced dataset along with the classifiers trainer in three rounds of boosting.}
\label{figMotiv}
\end{figure*}

\subsection{Literature on Boosting techniques tailored for handling class imbalance}

Several improvements over traditional boosting techniques have been proposed to tackle the problem of class imbalance. Some of the approaches integrate data balancing techniques with boosting \cite{galar2012}. For example, SMOTEBoost \cite{Chawla2003}, RAMOBoost \cite{chen2010}, DataBoost-IM \cite{guo2004learning}, and AMDO \cite{yang2017amdo} are amalgamations of minority class oversampling and boosting methods. On the other hand, JOUS-Boost \cite{mease2007cost} and RUSBoost \cite{seiffert2010rusboost} combine boosting with majority class undersampling. Wang and Yao \cite{wang2012multiclass} combined Negative Correlation based AdaBoost (AdaBoost.NC) with random oversampling of the minority classes to handle multi-class imbalanced problems.

There also exist methods which combine cost-sensitive learning (i.e. higher weight is assigned to the minority class) with boosting. The solutions range across techniques which incorporate cost-sensitivity into the update rule for the point-wise weights \cite{ting2000comparative, viola2002fast}, methods which modify the schemes for both point-wise weight update and component classifiers weights \cite{joshi2001evaluating, sun2007cost}, and techniques which induce cost-sensitivity into the error function and modify all aspects of boosting to comply with the modified error function \cite{masnadi2007asymmetric,masnadi2011cost}.

Another approach is to calibrate the ensemble learned by AdaBoost, post-training, to make the output scores of the ensemble correspond with class-probability estimates. Thereafter, the expected misclassification cost can be minimized by selecting the optimal threshold on the calibrated classifier scores, when the costs of misclassification for both the classes are known for a two-class problem. Nikolaou and Brown \cite{nikolaou2015calibrating} showed that a key advantage of this method, AdaBoost with Minimum Expected Cost and Calibration (AdaMEC-Calib), is the ability to account for changes in class imbalance without having to retrain the ensemble.

Based on the generally accepted notion that the minimum margin is a key to the generalization performance, Grove and Schuurmans \cite{grove1998boosting} presented an interesting variant of boosting, called LPAdaBoost, where the weights of the component classifiers obtained by AdaBoost are chosen by a Linear Program (LP) so as to maximize the minimum margin (margin can be thought of as the distance from the decision boundary in the direction of proper classification; see Section \ref{sec:towMotiv} for a formal definition). While LPAdaBoost is not aimed at handling class imbalance, it has innate ability to mitigate the effects of class imbalance. Since the component classifiers generated by AdaBoost are likely to be overwhelmed by the abundance of the majority class instances, most of the data points would tend to be placed into the majority class. This would result in high margins for the majority instances and low margins for the minority instances. Therefore, the minimum margin is likely to correspond to a minority instance. Consequently, LPAdaBoost, in its attempt to increase the minimum margin (corresponding to the most difficult minority point), would also improve the performance on the rest of the minority instances. However, in the presence of outliers, data noise or label noise, the minimum margin is likely to correspond to corrupt instances, leading to a complete miscalibration of the component weights. Leskovec and Shawe-Taylor \cite{leskovec2003linear} proposed LPUBoost, attempting to solve this problem by allowing regularization of the outliers and assigning higher regularization cost for the minority instances. 

It is important to understand that the performance of most of these boosting variants which are capable of tackling class imbalance require expensive cost set tuning to achieve optimal performance. For example, the proper extent of oversampling or undersampling for the sampling based techniques and the appropriate set of relative costs for the cost-sensitive techniques need to be determined by cost tuning. The best set of relative weights depends on a variety of factors such as the relative densities of the classes, the extent and structure of the overlap (if any) between the classes, the extent of noise, the number of outliers, etc. \cite{das2018handling}.

\subsection{Boosting as a game of \emph{Tug of War}}\label{sec:towMotiv}

A clear understanding of the effects of the component weights on the margin values can help us devise a cost-independent boosting method for handling class imbalance. Towards this end, we begin by formally defining margin in the context of boosting ensembles.

\begin{dfn}\label{dfn:margin}
Let $X = \{(\mathbf{x}_i,y_i): i = 1,2,\cdots,n; y_i \in \{-1,+1\}\}$ be a given training dataset and let $\{f_1, f_2, \cdots, f_T\}$ be the set of component classifiers having corresponding set of component weights $\{\alpha_1, \alpha_2, \cdots, \alpha_T\}$. Then for a data point $\mathbf{x}_i \in X$, the margin $\rho(\mathbf{x}_i)$ is defined as
\begin{equation*}
\begin{aligned}
& \rho(\mathbf{x}_i) = y_i h(\mathbf{x}_i),\\
\text{where } & h(\mathbf{x}_i) = \sum_{t=1}^{T} \alpha_t f_t(\mathbf{x}_i),\\
\end{aligned}
\end{equation*}
and is hereafter referred as the ``signed margin" of the point $\mathbf{x}_i$ while $f_t(\mathbf{x}_i)$ denotes the output of the $t$-th classifier for $\mathbf{x}_i$.
\end{dfn}

\begin{figure*}[h]
\vskip 0.2in
\begin{center}
\centerline{\includegraphics[width=0.97\textwidth]{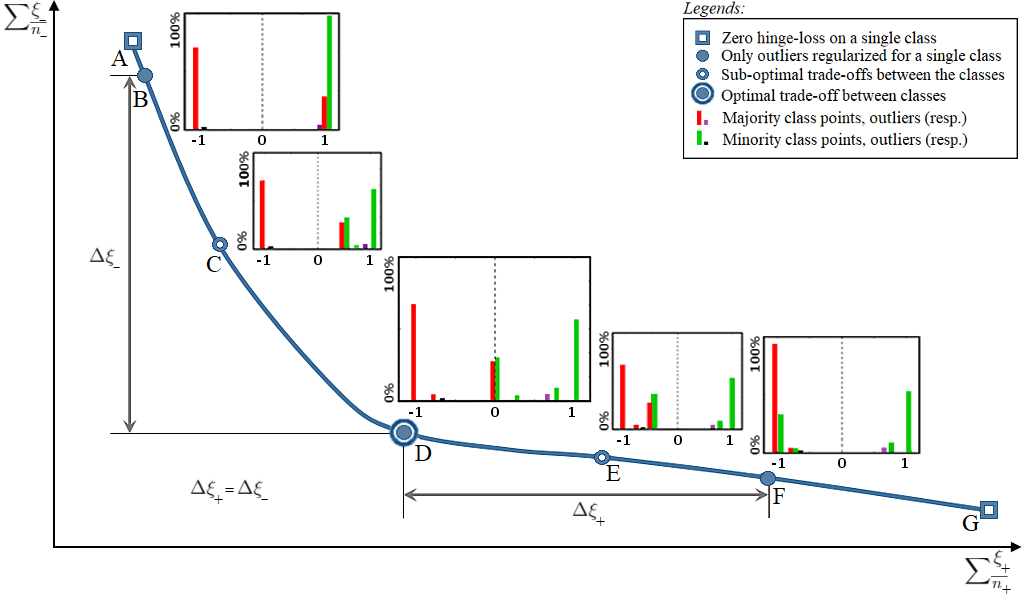}}
\vspace{-4mm}
\caption{\emph{Tug of War} between the classes: Solutions A and G respectively correspond to ideal performance on the minority and majority classes, and may not be attainable in practice on most datasets. B and F correspond to the respective best attainable performances with only the outliers having non-ideal signed margins. C, D, and E are trade-off solutions with non-outlier points from both classes having non-ideal signed margins. Solution D yields the optimal trade-off with similar fractions of non-outlier points having zero margin (i.e. prone to being misclassified) from both classes.}\label{fig:tugOfWar}
\end{center}
\vskip -0.2in
\end{figure*}

Let us consider the simple imbalanced dataset exhibited in Figure \ref{figMotivData}. It is seen that both the majority as well as the minority classes contain outliers. We run three rounds of AdaBoost on this dataset to obtain three component classifiers (see Figures \ref{figMotiv1}-\ref{figMotiv3}). 
Ideally, if the component classifiers output labels in the set $\{-1,+1\}$, each point in the positive class should have a signed margin value of +1 while all the negative points should have a signed margin value of -1 (i.e. all points should ideally have margin values of +1). An inspection of Figures \ref{figMotiv1}-\ref{figMotiv3} shows that some of the points from both the classes are always correctly classified. These points will always attain the ideal margin value, irrespective of the component weights. However, the margin values attained by other points, which are correctly classified only by some of the components, depends on the choice of component weights. If the components which correctly classify most of the majority class instances (generally at the cost of the minority class instances) are assigned high weightage, most of the majority points will have high margin values while many of the minority points will have low margin values. Similarly, if high weightage is assigned to classifiers performing well on the minority points, most minority points will have high margin values at the expense of having low margin values for the majority instances. Therefore, one can think of the problem of component weight assignment as a game of \emph{Tug of War}\footnotemark  in the margin space (in the sense increase in the margin values for minority points results in decrease in the margin values for majority points, and \emph{vice versa}).

\footnotetext{\emph{Tug of War} is formally defined in mathematics as a zero-sum two-player game where the losses for the two players add up to zero \cite{charro2009mixed, peres2011tug}. Our formulation is analogous to these formulations in the sense that the two classes can be considered as the two players and increase in the margin values for points in one class must come at the expense of decrease in the margin values for points from the opposite class. Yet, our formulation is distinct in that the net sum of changes in the margin values may not be zero.}

\subsection{\emph{Tug of War} for handling Class Imbalance}\label{sec:towMotiv2}

Intuitively, if both the classes are of equal importance or if no cost information is available, the best trade-off solution is to select the component weights so that similar fractions of data points are misclassified from both classes. A common way to quantify the extent of misclassification from the classes, employed by methods like LPBoost \cite{ratsch2001soft} and LPUBoost \cite{leskovec2003linear}, is to measure the average hinge losses (on the difference between the actual and ideal margin values) for the classes.
\begin{dfn}\label{dfn:hinge}
For a data instance $\mathbf{x}_i \in X$, the hinge loss on the difference $\rho(\mathbf{x}_i) - 1$ is defined as
\begin{equation*}
L_h(\mathbf{x}_i) =
\begin{cases}
0 \quad \quad \quad \;\; \text{ if } \rho(\mathbf{x}_i) >= 1, \\
1 - \rho(\mathbf{x}_i) \text{ if } \rho(\mathbf{x}_i) < 1.
\end{cases}
\end{equation*}
\end{dfn}
In the absence of outliers, similar average hinge losses from the two classes is likely to indicate similar fractions of misclassification. However, in the presence of outliers (due to noise or otherwise), the high hinge losses for the outlier instances can result in disproportionate increase in the average hinge losses, resulting in miscalibration. Such miscalibration leads to the failure of hard margin maximization \cite{ratsch2001soft} in the presence of outliers. Therefore, there is a need to regularize the outliers by some means. Therefore, we are motivated to devise a new framework, based on the \emph{Tug of War} analogy, which can effectively regularize the outlier instances. Consequently, this framework can be used to strike a good compromise between the two classes (in the sense of having similar fractions of misclassification of non-outlier instances), without resorting to cost set tuning.

The \emph{Tug of War} game for the example presented in Figure \ref{figMotiv} is illustrated in Figure \ref{fig:tugOfWar}. The solutions designated as A, and G in Figure \ref{fig:tugOfWar} respectively correspond to the cases where all points from the minority and majority classes attain the ideal margin value. In real-world applications, some difficult instances may be misclassified by all component classifiers. Hence, such ideal solutions are often unattainable in practice and may not correspond to any set of feasible component weights. The solutions B, and F correspond to the sets of component weights which minimize the individual average hinge losses respectively for the minority and majority classes. Being surrounded by points from the opposite class, outliers are generally correctly classified only after accumulating high weightage. However, at such high weights, the component classifiers are likely to misclassify many of the non-outlier points due to the influence of the borderline and/or outlier instances from the opposite class. Consequently, these components are generally assigned lower weightage while attempting to minimize the average hinge loss, resulting in the regularization of the outliers (in the sense that the outliers have worse signed margin and consequently higher loss). Now, any set of component weights which seeks to achieve a balance between the classes will result in an increase beyond the individual minimum average hinge losses for both the classes. Solutions C, D, and E in Figure \ref{fig:tugOfWar} correspond to such component weight combinations. Since the outliers already had high loss values, most of this increase in loss will be due to the misclassification of the non-outlier instances. Therefore, similar increase in the average hinge losses of the two classes will correspond to similar fraction of misclassification of non-outlier points for the two classes. Hence, the optimal trade-off between the classes can be achieved by finding the set of components weights corresponding to equal increase in the average hinge loss values for the minority and majority classes, w. r. t. the solutions B and F, respectively. Solution D in Figure \ref{fig:tugOfWar} corresponds to the optimal trade-off characterized by similar fractions of non-outlier points from the two classes having margin values close to zero.



\vspace{-2mm}
\subsection{Brief overview of literature on Lexicographic Programming}\label{sec:LxPlit}

In this paper, we show how the optimal trade-off among classes can be achieved by using Lexicographic Programming (LxP). LxP is concerned with solving a hierarchy of optimization problems where the objective function and/or the constraints imposed on a problem in the hierarchy depend on the optimal solutions obtained for one or more of the problems solved in prior stages of the hierarchy. A special case of LxP consists of solving a sequence (i.e. a hierarchy with exactly one problem in each stage) of optimization problems. This framework is generally employed for Multi-Criteria Decision Making (MCDM) in the forms of Lexicographic Goal Programming (LGP) and Lexicographic Multi-Objective Programming (LMOP) \cite{ignizio1976goal,SANKARAN1998669}. LGP attempts to attain predefined goals for a set of objectives which are arranged in decreasing order of priority. LMOP, on the other hand, aims to minimize the set of objectives in order of priority. Due to ease of solving LP problems, LGP and LMOP problems are traditionally formulated as a sequence of LPs \cite{ijiri1965management,lee1972goal,ignizio1976goal}. There have also been efforts to reduce LGP and LMOP problems to single objective optimization problems using various approaches \cite{SANKARAN1998669,POURKARIMI20071348,COCOCCIONI2018298}. However, such reductions are generally not applicable to the general LxP problems consisting of multiple optimization problems in each stage of the hierarchy. Romero \cite{ROMERO200163} showed that a large number of MCDM problems can be shown to be equivalent to a general formulation of LGP.

\subsection{Contributions}\label{sec:contrib}

Class imbalanced classification can also be thought of as a multi-criteria decision making problem, since the classification accuracy on the majority as well as the minority classes must be simultaneously maximized (these two objectives are often contradictory and cannot be maximized together, resulting in the need for a suitable trade-off). In spite of this, to the best of our knowledge, the current article is the first application of LxP to the class imbalanced classification problem.

\begin{dfn}
We formally define a Lexicographic Linear Program (LxLP) as a lexicographic hierarchy of LPs (in the sense that the LPs in all the prior stages of the hierarchy must be solved before the LPs in the current stage can be solved). The $j$-th LP to be solved at the $i$-th stage is of the form
\begin{equation*}
\begin{aligned}
& \mathbf{v}_{ij}^{*} = \argmin_{\mathbf{v}} \;\; L_{ij}(\mathbf{v},\mathbf{v}_{i-1}^{*}), \\
\text{s. t. } & g_k(\mathbf{v},\mathbf{v}_{i-1}^{*}) \leq 0 \text{ }\forall k \in \{1,\cdots,\eta_i\} \\
\text{and } & h_l(\mathbf{v},\mathbf{v}_{i-1}^{*})=0 \text{ }\forall l \in \{1,\cdots,\nu_i\},
\end{aligned}
\end{equation*}
where $\eta_i$ and $\nu_i$ respectively are the number of inequality and equality constraints while $L_{ij}$ is an appropriate loss function. The vector $\mathbf{v}_{i-1}^{*} = (\mathbf{v}_{(i-1)1}^{*},\cdots,\mathbf{v}_{(i-1)\zeta_{i-1}}^{*},\cdots,\mathbf{v}_{11}^{*},\cdots,\mathbf{v}_{1 \zeta_1}^{*})$ contains the optimal solutions to all LPs solved in all the preceding stages with $\zeta_{i}$ denoting the number of LPs solved in the $i$-th stage.
\end{dfn}

Based on the insights from Section \ref{sec:towMotiv2}, we propose a two staged scheme to choose the weights of the component classifiers of a boosted ensemble. The first stage is concerned with solving a set of LPs (one for each class, which can be solved in parallel) to find the two (possibly different) sets of component weights corresponding to the individual minimum attainable average hinge losses for the two classes. Subsequently, the second stage solves another LP to find the set of component weights that minimizes the maximum increase in the class-wise average hinge losses beyond the minimum values found in the first stage. It is clear that the proposed scheme is an LxLP problem as all LPs in the first stage must be solved before the final LP from the second stage can be formulated. The proposed method is referred to as LexiBoost hereafter. We also formulate a dual to LexiBoost, called Dual-LexiBoost, which not only selects optimal component weights but also adapts the point-wise weights, over the rounds of boosting, to counter the effects of class imbalance.

The proposed methods have the following advantages:
\begin{itemize}
\item It invokes the novel hinge loss based regularization method, which unlike the slack variable based regularization, does not require cost set tuning to achieve a good balance between the classes, thus addressing the long-standing issue of expensive cost set tuning for imbalanced data learning.
\item Moreover, the proposed approach is readily applicable to multi-class or polychotomous learning tasks, which have as yet received limited attention in the class imbalanced learning literature.
\item Even though we demonstrate the abilities of the proposal by using the AdaBoost algorithm, the proposed philosophy can be readily applied to other ensemble learning techniques (such as bagging \cite{breiman1996bagging}) as well.
\end{itemize}

\subsection{Organization}\label{sec:org}

We introduce the reader to some of the existing LP based boosting schemes in Section \ref{sec:lp}. We then provide a detailed explanation of the proposed two staged LxLP based LexiBoost framework in Section \ref{sec:lexiboost}. The dual formulation resulting from the proposed LxLP is presented in Section \ref{sec:duallexi}. The proposed framework is also generalized to multi-class classification problems in Section \ref{sec:multiClass}. Computational complexity of the proposed methods is discussed in Section \ref{sec:timeComp}. Subsequently, experimental results are presented and discussed in Section \ref{sec:exp}. We conclude the article in Section \ref{sec:concl}.

\section{Linear Programming based Boosting}\label{sec:lp}

In this section, we introduce the reader to some of the extant LP based boosting techniques which are crucial to understanding the proposed improvement.

\textbf{LPAdaBoost:} Despite the theoretical guarantees on the training performance of AdaBoost \cite{freund1995desicion}, Grove and Schuurmans \cite{grove1998boosting} proposed the LPAdaBoost algorithm to maximize the minimum margin $\rho$, aiming to achieve better generalization performance. The primal LP posed by LPAdaBoost is of the form
\begin{equation*}
\begin{aligned}
\text{A}_1\text{: } & (\bm{\alpha}^{*},\rho^{*}) = \argmax \;\; \rho, \\
\text{s. t. } & y_i \sum_{t=1}^{T} \alpha_t f_t(\mathbf{x}_i) \geq \rho \text{ } \forall i \in \{1,\cdots,n\}, \\
& \sum_{t=1}^{T} \alpha_t = 1, \\
\text{and } & \bm{\alpha} \geq 0,
\end{aligned}
\end{equation*}
where $f_t(\mathbf{x}_i)$ is the classifier generated by the $t$-th round of AdaBoost.

\textbf{Dual-LPAdaBoost:} The dual to the primal LPAdaBoost formulation is
\begin{equation*}
\begin{aligned}
\text{A}'_1\text{: } & (\bm{D}_{t+1}^{*},s^{*}) = \argmin \;\; s, \\
\text{s. t. } & \sum_{i=1}^{n} D_{t+1}(i) y_i f'_{\tau}(\mathbf{x}_i) \leq s \text{ } \forall \tau \in \{1,\cdots,t\}, \\
& \sum_{i=1}^{n} D_{t+1}(i) = 1, \\
\text{and } & \bm{D}_{t+1} \geq 0,
\end{aligned}
\end{equation*}
where $f'_{\tau}(\mathbf{x}_i)$ is the classifier generated in the $\tau$-th round. The dual formulation corresponds to assigning the point-wise weights $D_{t+1}(i)$ such that the aggregate margin $s$ is minimized. To put it simply, the dual attempts to find a $\bm{D}_{t+1}$ which assigns the greatest weightage to the points which prove to be the most difficult during rounds $1$ through $t$. The Dual-LPAdaBoost algorithm consists of alternatingly solving the LPs A$_1$ and A$'_1$ until the convergence criterion $s - \rho < 0$ is met or the maximum number of rounds $T$ is reached. 

\textbf{LPBoost:} Since the hard margin formulation of LPAdaBoost makes it sensitive to noise and outliers, R{\"a}tsch et al. \cite{ratsch2001soft} presented a soft margin variant called LPBoost which regularizes the outlier instances using slack variables $\xi_i$ (corresponding to the data points $\mathbf{x}_i$), resulting in the following LP:
\begin{equation*}
\begin{aligned}
\text{A}_2\text{: } & (\bm{\alpha}^{*},\bm{\xi}^{*},\rho^{*}) = \argmin \;\; - \rho + D \sum_{i=1}^{n} \xi_i, \\
\text{s. t. } & y_i \sum_{t=1}^{T} \alpha_t f_t(\mathbf{x}_i) \geq \rho - \xi_i \text{ } \forall i \in \{1,\cdots,n\}, \\
& \sum_{t=1}^{T} \alpha_t = 1, \\
& \bm{\alpha} \geq 0, \\
\text{and } & \bm{\xi} \geq 0.
\end{aligned}
\end{equation*}
An appropriately high cost $D$ must be assigned for the regularization of data points in order to achieve good performance. This parameter has to be generally selected by cross-validation on the training data. 

\textbf{DualLPBoost:} The dual LP arising out of the LPBoost formulation is of the form
\begin{equation*}
\begin{aligned}
\text{A}'_2\text{: } & (\bm{D}_{t+1}^{*},s^{*}) = \argmin \;\; s, \\
\text{s. t. } & \sum_{i=1}^{n} D_{t+1}(i) y_i f_{\tau}(\mathbf{x}_i) \leq s \text{ } \forall \tau \in \{1,\cdots,t\}, \\
& \sum_{i=1}^{n} D_{t+1}(i) = 1, \\
\text{and } & 0 \leq \bm{D}_{t+1} \leq D,
\end{aligned}
\end{equation*}
giving rise to the Dual-LPBoost algorithm where A$'_2$ is solved for a maximum of $T$ rounds (until the convergence criterion $\sum_{i=1}^{n} D_t(i) y_i f_t(\mathbf{x}_i) \leq s$ is satisfied) with the Lagrangian multipliers of A$'_2$ being chosen to be the component weights $\alpha_t$.

\textbf{LPUBoost:} Leskovec and Shawe-Taylor \cite{leskovec2003linear} further adapted the LPBoost formulation to two-class imbalanced problems by introducing uneven costs for regularizing the two classes. The non-target (usually majority) class is assigned a regularization cost of $D$ as in LPBoost, while the target (usually minority) class is assigned a higher regularization cost of $\beta D$ $(\beta > 1)$. The resulting primal LP is
\begin{equation*}
\begin{aligned}
\text{A}_3\text{: } & (\bm{\alpha}^{*},\bm{\xi}^{*},\rho^{*}) = \argmin \;\; - \rho + \beta D \sum_{i=1}^{n_{1}} \xi_i + D \sum_{i=n_1+1}^{n} \xi_i, \\
\text{s. t. } & y_i \sum_{t=1}^{T} \alpha_t f_t(\mathbf{x}_i) \geq \rho - \xi_i \text{ } \forall i \in \{1,\cdots,n\}, \\
& \sum_{t=1}^{T} \alpha_t = 1, \\
\text{and } & (\bm{\alpha},\bm{\xi}) \geq 0,
\end{aligned}
\end{equation*}
where $n_{1}$ denotes the number of points in the positive (target) class. Hence, the number of points in the negative (non-target) class is $n_{2} = n - n_1$. Both the parameters $D$ as well as $\beta$ must be selected by expensive tuning on $\mathbb{R}^{+} \times \mathbb{R}^{+}$ using cross-validation. Tuning the parameter $\beta$ essentially corresponds to tuning the relative cost between the two classes, and is critical to achieving good performance.

\textbf{Dual-LPUBoost:} Like LPAdaBoost and LPBoost, LPUBoost also gives rise to a dual problem. The dual problem is of the form
\begin{align}
\text{A}'_3\text{: } & (\bm{D}_{t+1}^{*},s^{*}) = \argmin \;\; s, \nonumber \\
\text{s. t. } & \sum_{i=1}^{n} D_{t+1}(i) y_i f_{\tau}(\mathbf{x}_i) \leq s \text{ } \forall \tau \in \{1,\cdots,t\}, \nonumber \\
& \sum_{i=1}^{n} D_{t+1}(i) = 1, \nonumber \\
& 0 \leq D_{t+1}(i) \leq \beta D \text{ } \forall i \in \{1,\cdots,n_1\}, \label{constr1} \\
\text{and } & 0 \leq D_{t+1}(i) \leq D \text{ } \forall i \in \{n_1+1,\cdots,n\}. \label{constr2}
\end{align}
However, the Dual-LPUBoost algorithm solves a slightly modified from of the LP A$'_3$ to accommodate for the drawbacks of simple cost set tuning. While the modification does seem to lend some robustness to the method (see Section \ref{sec:exp}), it also adds an additional tunable parameter $D_{LB}$, which determines the lower limit of $D_{t+1}(i)$ as a fraction of the corresponding upper limit. Thus, the changes pertain to the constraints (\ref{constr1}) and (\ref{constr2}), resulting in the modified constraints
\begin{equation*}
\begin{aligned}
& \beta D \times D_{LB} \leq D_{t+1}(i) \leq \beta D \text{ } \forall i \in \{1,\cdots,n_1\}, \\
\text{and } & D \times D_{LB} \leq D_{t+1}(i) \leq D \text{ } \forall i \in \{n_1+1,\cdots,n\}.
\end{aligned}
\end{equation*}
The termination criterion and the choice of component weights are identical to those of Dual-LPBoost.


\section{Lexicographic Linear Programming based selection of component weights}\label{sec:lexiboost}

Having acquainted the reader to the existing LP based boosting schemes, we now elucidate the proposed LexiBoost algorithm which uses a two staged LxLP. The two stages of LP involved in our proposed LxLP framework are formally defined in Sections \ref{sec:initOpt} and \ref{sec:finalOpt}.

\subsection{The first stage of LPs}\label{sec:initOpt}

As already mentioned in Section \ref{sec:contrib}, the initial aim is to choose the component classifier weights so that the average hinge loss on the differences between the actual and ideal margins is minimized for the individual classes. Since the hinge loss is piece-wise linear, the minimization problems can be posed as LPs. Therefore, an LP P$_j$ ($j$ denotes the class in question) of the following form must be solved for each of the classes:
\begin{align}
\text{P}_j\text{: } & (\bm{\alpha}_j,\bm{\lambda}_{j}^{*}) = \argmin \;\; \frac{1}{n_j} \sum_{i=1}^{n_j} \lambda_i, \nonumber \\
\text{s. t. } & 1 - \rho(\mathbf{x}_i) \leq \lambda_i \text{ } \forall i \in \{1,\cdots,n_j\}, \label{Pconstr1} \\
& \sum_{t=1}^{T} \alpha_t = 1, \label{Pconstr2} \\
& \bm{\alpha} \geq 0, \label{Pconstr3} \\
\text{and } & \bm{\lambda} \geq 0, \label{Pconstr4}
\end{align}
where $n_j$ is the number of points in the $j$-th class ($j \in \{1,\cdots,|\mathcal{C}|\}$, $\mathcal{C} = \{c_1,c_2\}$ for the two-class imbalanced problem being the set of classes), $\lambda_i$ are the slack variables which measure the hinge loss for the points $\mathbf{x}_i$ by serving as the upper bound on the difference between the ideal and actual margin values (which depends on the choice of the component weights $\alpha_t$ as per Definition \ref{dfn:margin}). $\bm{\lambda}_{j}^{*}$ denotes the optimal set of $\lambda_i$ values obtained by solving the LP P$_j$, and $\bm{\alpha}_j$ is the corresponding set of component weights.

\subsection{The final LP}\label{sec:finalOpt}

Since the outlier instances already have high loss values even when the overall class-wise losses are minimized, any further increase in the class-wise losses is likely to be due to regularization of non-outlier instances. Therefore, having obtained the optimal hinge losses $\bm{\lambda}^*_j$ for each of the individual classes, the deviations from average optimal losses must be minimized to restrict the regularization (and hence misclassification) of non-outlier points. Hence, we finally solve another LP, referred to as Q, to find the $\alpha_t$ values which minimize $\chi$ the maximum of such deviations as per constraint (\ref{Qconstr1}). $\chi^{*}$ denotes the optimized value of $\chi$. It should be noted that unlike $P_j$, all the classes (and hence all the training data points) are considered together at this stage. The LP Q is run to obtain the optimal set of component weights $\bm{\alpha}^{*}$ which strike a good balance between the classes. The formulation\footnotemark \, for the program is as follows:
\begin{align}
\text{Q: } & (\bm{\alpha}^{*},\bm{\lambda},\chi^{*}) = \argmin \;\; \chi, \nonumber \\
\text{s. t. } & \frac{1}{n_j} \sum_{i \in c_j} \lambda_i - \frac{1}{n_j} \mathbf{e}^T \bm{\lambda}_{j}^{*} \leq \chi \text{ } \forall j \in \{1,\cdots,|\mathcal{C}|\}, \label{Qconstr1} \\
& 1 - \rho(\mathbf{x}_i) \leq \lambda_i \text{ } \forall i \in \{1,\cdots,n\}, \label{Qconstr2} \\
& \sum_{t=1}^{T} \alpha_t = 1, \label{Qconstr3} \\
& \bm{\alpha} \geq 0, \label{Qconstr4} \\
& \bm{\lambda} \geq 0, \label{Qconstr5} \\
\text{and } & \chi \geq 0, \label{Qconstr6}
\end{align}
where $\lambda_i$ denotes the new loss (possibly different from the loss $\lambda^*_{ji} \in \bm{\lambda}^*_j$ obtained from P$_j$) for the point $\mathbf{x}_i \in c_j$, $\alpha_t$ are the corresponding component weights, and $\mathbf{e}$ denotes a vector of ones, having appropriate length. Hence, the complete LexiBoost algorithm (in its primal formulation) is presented in Algorithm 1.

\footnotetext{If a known cost $\mathfrak{C}_j$ is attached to the class $c_j$ in an application, constraint (\ref{Qconstr1}) should be modified to $\frac{\mathfrak{C}_j}{n_j} \sum_{i \in c_j} \lambda_i - \frac{\mathfrak{C}_j}{n_j} \mathbf{e}^T \bm{\lambda}_{j}^{*} \leq \chi \text{ } \forall j \in \{1,\cdots,|\mathcal{C}|\}$.}

\begin{figure}[!ht]
\begin{algorithmic}
   \State \hrulefill
   \State \textbf{Algorithm 1:} LexiBoost
   \State \hrulefill
   \State \textbf{Input:} Dataset $X$ of labelled instances $\mathbf{x}_i$.
   \State \textbf{Output:} Final ensemble classifier\\
   $\;\;\;\;\;\;\;\;\;\;\;\;\;\;\;\;\;\;\;\;$ $H(\mathbf{x}_i) = \text{sign}(\sum_{t=1}^{T} \alpha_t^{*} f_t(\mathbf{x}_i))$.
   \State \hrulefill
   \State Run AdaBoost on the dataset $X$ to obtain $f_t(\mathbf{x}_i)$.
   \ForAll{$j \in \{1,\cdots,|\mathcal{C}|\}$}
        \State Solve LP P$_j$ to obtain $\bm{\lambda}_j^{*}$.
   \EndFor
   \State Solve LP Q to obtain $\bm{\alpha}^{*}$.
   \State \hrulefill
\end{algorithmic}
\vspace{-5mm}
\end{figure}

\subsection{The dual to LexiBoost}\label{sec:duallexi}

The dual to boosting techniques are often of interest. This is principally because the dual formulations allow not only for suitable choice of the component weights but also facilitate suitable choices of the point-specific weights over the rounds. Moreover, from a theoretical point of view, the dual formulations can be used to unify seemingly different boosting techniques such AdaBoost, LPBoost, etc. \cite{onDual2010}. Therefore, in this section we present the dual formulation of the proposed LexiBoost algorithm. 

The Lagrangians arising from the LPs P$_j$ of the LexiBoost primal are of the form
\begin{equation*}
\begin{aligned}
    \mathcal{L}_j(\bm{\alpha},\bm{\lambda},\bm{D},\bm{\sigma},\bm{\gamma},s) & = \frac{1}{n_j} \sum_{i=1}^{n_j} \lambda_i + s(\sum_{t=1}^{T} \alpha_t - 1)\\ 
    + \sum_{i=1}^{n_j} D(i) (1 - \lambda_i & - y_i \sum_{t=1}^{T} \alpha_t f_t (\mathbf{x}_i)) - \sum_{t=1}^{T} \sigma_t \alpha_t - \sum_{i=1}^{n_j} \gamma_i \lambda_i,
\end{aligned}
\end{equation*}
where $D(i)$, $s$, $\sigma_t$, and $\gamma_i$ are respectively the Lagrangian multipliers corresponding to the constraints (\ref{Pconstr1}), (\ref{Pconstr2}), (\ref{Pconstr3}), and (\ref{Pconstr4}). Differentiating $\mathcal{L}_j$ w. r. t. $\lambda_i$ and $\alpha_t$ and equating to zero, we get the constraints
\begin{align}
    & \frac{\partial \mathcal{L}_j}{\partial \lambda_i} = \frac{1}{n_j} - D(i) - \gamma_i = 0, \label{PdualConstr1} \\
    \text{and } & \frac{\partial \mathcal{L}_j}{\partial \alpha_t} = s - \sum_{i=1}^{n_j} D(i) y_i f_t(\mathbf{x}_i) - \sigma_t = 0. \label{PdualConstr2}
\end{align}
Imposing constraints (\ref{PdualConstr1}) and (\ref{PdualConstr2}) in the Lagrangian, and eliminating $\sigma_t$ and $\gamma_i$ from the constraints, we get the dual LPs of the form
\begin{equation}\label{eqn:initDual}
\begin{aligned}
& \text{maximize } \sum_{i=1}^{n_j} D(i) - s, \\
\text{s. t. } & \sum_{i=1}^{n} D(i) y_i f_t (\mathbf{x}_i) \leq s, \\
& 0 \leq D(i) \leq \frac{1}{n_j} \text{ } \forall i \in c_j.
\end{aligned}
\end{equation}
Now, since the dual LPs for boosting algorithms can be used for choosing suitable point-specific weights subsequent to each round of boosting, we slightly modify the LP in (\ref{eqn:initDual}) to obtain the dual LPs P'$_j$ of the form
\begin{align}
\text{P}'_j\text{: } & (\bm{D}_{t+1}^{*},s^{*}) = \argmax \;\; \sum_{i=1}^{n_j} D_{t+1}(i) - s, \nonumber\\
\text{s. t. } & \sum_{i=1}^{n} D_{t+1}(i) y_i f_{\tau}(\mathbf{x}_i) \leq s \text{ } \forall \tau \in \{1,\cdots,t\}, \nonumber\\
& \sum_{i=1}^{n} D_{t+1}(i) = 1, \label{PconstrSum} \\
& 0 \leq D_{t+1}(i) \leq \frac{1}{n_j} \text{ } \forall i \in c_j, \forall j \in \{1,\cdots,|\mathcal{C}|\}, \label{PconstrBounds}
\end{align}
where $D_{t+1}(i)$ (corresponding to the $D(i)$ in (\ref{eqn:initDual})) is the point-specific weight of the data point $\mathbf{x}_i$. We have imposed the additional constraint (\ref{PconstrSum}) to ensure that the $D_{t+1}(i)$ values can be meaningfully used as data point weights. It is easy to see that the LP remains feasible despite the introduction of constraint (\ref{PconstrSum}) as the case where $D_{t+1}(i)=\frac{1}{|\mathcal{C}| n_j}$ $\forall i \in c_j$ $\forall j \in \{1,\cdots,|\mathcal{C}|\}$ is a feasible solution. However, the introduction of the constraints means that a sub-optimal solution for (\ref{eqn:initDual}) may be the optimal solution for P$'_j$, resulting in the optimal objective function value of P$'_j$ being lower than that of P$_j$ (despite there being no duality gap between (\ref{eqn:initDual}) and P$_j$, as strong duality holds for LP problems).

The interpretation of P$'_j$ is similar to that of LPAdaBoost in that the dual formulation attempts to assign higher weights to the points which have proved to be difficult in the previous rounds $1$ through $t$ while also maximizing the sum of weights. The constraints (\ref{PconstrBounds}) ensure that higher weights are assigned to instances belonging to the minority class.

Additionally, the Lagrangian corresponding to the final LP Q of the LexiBoost primal is as follows:
\vspace{-2mm}
\begin{equation*}
\begin{aligned}
    \mathcal{L}(\bm{\alpha},\bm{\lambda},\chi,\bm{d},\bm{D},\bm{\sigma},\bm{\gamma},s,\psi) = \chi + s(\sum_{t=1}^{T} \alpha_t - 1) \\
    \sum_{j=1}^{|\mathcal{C}|} d_j (\frac{1}{n_j} \sum_{i \in c_j} \lambda_i - \frac{1}{n_j} \sum_{i \in c_j} \lambda_{ji}^{*} - \chi) \\
    + \sum_{i=1}^{n_j} D(i) (1 - \lambda_i - y_i \sum_{t=1}^{T} \alpha_t f_t (\mathbf{x}_i)) \\ - \sum_{t=1}^{T} \sigma_t \alpha_t - \sum_{i=1}^{n_j} \gamma_i \lambda_i - \psi \chi,
\end{aligned}
\end{equation*}
where $d_j$, $D(i)$, $s$, $\sigma_t$, $\gamma_i$, and $\psi$ are respectively the Lagrangian multipliers corresponding to the constraints (\ref{Qconstr1})-(\ref{Qconstr6}). Differentiating $\mathcal{L}$ w. r. t. $\chi$, $\lambda_i$, and $\alpha_t$ and equating to zero, we get the constraints
\begin{align}
    & \frac{\partial \mathcal{L}_j}{\partial \chi} = 1 - \sum_{j=1}^{|\mathcal{C}|} d_j - \psi = 0, \label{QdualConstr1} \\
    & \frac{\partial \mathcal{L}_j}{\partial \lambda_i} = \frac{d_j}{n_j} - D(i) - \gamma_i = 0, \label{QdualConstr2} \\
    \text{and } & \frac{\partial \mathcal{L}_j}{\partial \alpha_t} = s - \sum_{i=1}^{n_j} D(i) y_i f_t(\mathbf{x}_i) - \sigma_t = 0. \label{QdualConstr3}
\end{align}
Imposing constraints (\ref{QdualConstr1})-(\ref{QdualConstr3}) in the Lagrangian, and eliminating $\sigma_t$, $\gamma_i$, and $\psi$ from the constraints, we get the dual LPs of the form
\begin{align}
& \text{max. } \sum_{i=1}^{n} D(i) - \sum_{j=1}^{|\mathcal{C}|} \frac{d_j}{n_j} \mathbf{e}^T \bm{\lambda}_{j}^{*} - s, \nonumber \\
\text{s. t. } & \sum_{i=1}^{n} D(i) y_i f_t(\mathbf{x}_i) \leq s, \nonumber \\
& 0 \leq D(i) \leq \frac{d_j}{n_j} \text{ } \forall i \in c_j, \forall j \in \{1,\cdots,|\mathcal{C}|\}, \nonumber \\
& \sum_{j=1}^{|\mathcal{C}|} d_j \leq 1, \text{ } d_j \geq 0, \nonumber
\end{align}
which we modify in a manner similar to the LPs P$'_j$ to obtain the following LP Q$'$:
\begin{align}
\text{Q}'\text{: } & (\bm{D}_{t+1}^{*},\bm{d},s^{*}) = \argmax \;\; \sum_{i=1}^{n} D_{t+1}(i) - \sum_{j=1}^{|\mathcal{C}|} \frac{d_j}{n_j} \mathbf{e}^T \bm{\lambda}_{j}^{*} - s, \nonumber \\
\text{s. t. } & \sum_{i=1}^{n} D_{t+1}(i) y_i f_{\tau}(\mathbf{x}_i) \leq s \text{ } \forall \tau \in \{1,\cdots,t\}, \nonumber \\
& \sum_{i=1}^{n} D_{t+1}(i) = 1, \label{QconstrSum} \\
& 0 \leq D_{t+1}(i) \leq \frac{d_j}{n_j} \text{ } \forall i \in c_j, \forall j \in \{1,\cdots,|\mathcal{C}|\}, \nonumber \\
& \sum_{j=1}^{|\mathcal{C}|} d_j \leq 1, \nonumber \\
& \bm{d} \geq 0. \nonumber
\end{align}
It is easy to see that the introduction of the constraint (\ref{QconstrSum}) does not make Q$'$ infeasible as ensuring $D_{t+1}(i)=\frac{d_j}{n_j}$ $\forall i \in c_j$ $\forall j \in \{1,\cdots,|\mathcal{C}|\}$ for some $\bm{d}$ such that $\sum_{j=1}^{|\mathcal{C}|} d_j = 1$ is enough to yield feasible solutions. However, similar to P$'_j$, the introduction of the additional constraint may result in the optimal objective function value of Q$'$ being lower than that of Q.

It can be seen that the LP Q$'$ is similar to the LPs P$'_j$ and only differs in that the upper bounds of the instance weights $D_{t+1}(i)$ are scaled by an amount $d_j$ (for $\mathbf{x}_i \in c_j$) which is inversely proportional to the average hinge loss obtained for the corresponding class in the first stage of LPs. In other words, greater regularization is induced (by enforcing a lower upper bound for instance weights) for the class having a greater proportion of outlier instances, while also maintaining higher weightage for the non-outlier minority class instances to compensate for class imbalance.

Based on these two dual formulations, we now present the complete Dual-LexiBoost method as Algorithm 2.
\begin{figure}[!ht]
\begin{algorithmic}
   \State \hrulefill
   \State \textbf{Algorithm 2:} Dual-LexiBoost
   \State \hrulefill
   \State \textbf{Input:} Dataset $X$ of labelled instances $\mathbf{x}_i$.
   \State \textbf{Output:} Final ensemble classifier\\
   $\;\;\;\;\;\;\;\;\;\;\;\;\;\;\;\;\;\;\;\;$ $H(\mathbf{x}_i) = \text{sign}(\sum_{t=1}^{T} \alpha_t^{*} f_t(\mathbf{x}_i))$.
   \State \hrulefill
   \State Initialize $D_1(i) = 1/(|\mathcal{C}| n_j)$ $\forall i \in c_j, \forall j \in \{1,\cdots,|\mathcal{C}|\}$.
   \ForAll{$t = 1$ to $T$}
        \State Train weak classifier $f_t$ using distribution $D_t$.
        \State Calculate training error $\epsilon_t$ using distribution $D_t$.
        \If{$\epsilon_t > \frac{1}{|\mathcal{C}|}$} break \EndIf
        	\ForAll{$j \in \{1,\cdots,|\mathcal{C}|\}$}
        			\State Solve LP P$'_j$ to obtain $D_{t+1}(i)$ $\forall i \in c_j$.
        	\EndFor
   \EndFor
   \ForAll{$j \in \mathcal{C}$}
        \State Solve LP P$_j$ to obtain $\bm{\lambda}_j^{*}$.
   \EndFor
   \State Initialize $D_1(i) = 1/(|\mathcal{C}| n_j)$ $\forall i \in c_j, \forall j \in \{1,\cdots,|\mathcal{C}|\}$.
   \ForAll{$t = 1$ to $T$}
        \State Train weak classifier $f_t$ using distribution $D_t$.
        \State Calculate training error $\epsilon_t$ using distribution $D_t$.
        \If{$\epsilon_t > \frac{1}{|\mathcal{C}|}$} break \EndIf
        \State Solve LP Q$'$ to obtain $D_{t+1}(i)$ $\forall i \in \{1,\cdots,n\}$.
   \EndFor
   \State Solve LP Q to obtain $\bm{\alpha}^{*}$.
   \State \hrulefill
\end{algorithmic}
\end{figure}

\subsection{Generalization to multi-class tasks}\label{sec:multiClass}

It is easy to see that the proposed approach is readily applicable to polychotomous or multi-class classification tasks. However, the definition of margin must be altered for the multi-class setting in the following way:
\begin{dfn}
Let $X = \{(\mathbf{x}_i, \mathbf{y}_i)\}$ be a given multi-class training dataset with label vectors $\mathbf{y}_i \in \mathbb{R}^{|\mathcal{C}|}$ such that
\begin{equation*}
y_{i,j} =
\begin{cases}
1 \text{ if $\mathbf{x}_i$ belongs to the $j$-th class}, \\
-1 \text{ otherwise}. \\
\end{cases}
\end{equation*}
Additionally, let $\mathbf{f}_t(\mathbf{x}_i)$ denote the prediction vector for the point $\mathbf{x}_i$ by the classifier $f_t$. Then the margin $\rho(\mathbf{x}_i)$ can be redefined as
\begin{equation*}
\rho(\mathbf{x}_i) = \sum_{t=1}^{T} \alpha_t {\mathbf{y}_i}^T \mathbf{f}_t(\mathbf{x}_i).
\end{equation*}
\end{dfn}
Thereafter, LexiBoost as well as Dual-LexiBoost can be directly applied to multi-class problems by modifying the LPs P$_j$, Q, P$'_j$, and Q$'$ accordingly. 

\subsection{Time Complexity of LexiBoost}\label{sec:timeComp}

It is well-known that the time complexity of interior-point methods for solving LPs is $O(\mathfrak{N}^3\mathfrak{L})$, where $\mathfrak{N}$ is the number of variables and $\mathfrak{L}$ is the size of the input data, i.e. the number of bits required to encode the coefficients of the objective function and the constraints of the LPs \cite{potra2000interior}. Now, the LPs P$_j$ of the LexiBoost primal have $(n_j + T)$ variables and requires $(2n_j + T + 1)(n_j + T + 1) + T$ coefficients for characterizing the objective function and the constraints. Assuming $T$ to be a constant and assuming that all the coefficients are encoded using a fixed number of bits, the time complexity of the LPs P$_j$ becomes $O(\sum_{j=1}^{|\mathcal{C}|} n_j^5)$. The final LP Q of the LexiBoost primal has $(n + T + 1)$ variables and is characterized by $(|\mathcal{C}| + 2n + T + 2)(n + T + 2) + T$ coefficients. Making the additional assumption that the number of classes is much lower than the number of data points (i.e. $n \gg |\mathcal{C}|$), the time complexity for Q becomes $O(n^5)$. Thus, the total complexity of the LexiBoost primal LPs becomes $O(n^5)$. On the other hand, the LPs P$'_j$ for Dual-LexiBoost have $(n + 1)$ variables and require $2(n + 1) + (t + n)(n + 2)$ coefficients, resulting in a complexity of $O(n^5)$. The final LP Q$'$ for Dual-LexiBoost has $(n + |\mathcal{C}| + 1)$ variables and is characterized by $(2n + |\mathcal{C}| + 2) + (t + n)(n + |\mathcal{C}| + 1)$ coefficients, resulting in a time complexity of $O(n^5)$. Hence, the LPs for Dual-LexiBoost have a total time complexity of $O(n^5)$. We compare the asymptotic order of time complexities of the methods discussed in Section \ref{sec:lp} against those of LexiBoost and Dual-LexiBoost in Table \ref{tab:times}. It can be observed from Table \ref{tab:times} that both LexiBoost and Dual-LexiBoost enjoy lower complexity compared to the primal as well as dual variants of LPBoost and LPUBoost as the proposed methods do not require cost tuning. For a comparison of the actual training times, see Figure \ref{figkNN}.

\begin{table}[!th]
\caption{Time complexities of Linear Programming based Boosting variants}\label{tab:times}
\begin{center}
\smallskip
\begin{tabular}{c|c|c}
\hline
Algorithm & Primal & Dual \\
\hline
LPAdaBoost & $O(n)$ & $O(n^5)$ \\
LPBoost & $O(C_D n^5)$ & $O(C_D n^5)$ \\
LPUBoost & $O(C_{\beta} C_D n^5)$ & $O(C_{\beta} C_D C_{LB} n^5)$ \\
LexiBoost & $O(n^5)$ & $O(n^5)$ \\
\hline
\multicolumn{3}{p{.34\textwidth}}{$C_D$: No. of candidate values of the parameter $D$ for LPBoost and Dual-LPBoost.}\\
\multicolumn{3}{p{.34\textwidth}}{$C_{\beta}$: No. of candidate values of the parameter $\beta$ for LPUBoost and Dual-LPUBoost.}\\
\multicolumn{3}{p{.34\textwidth}}{$C_{LB}$: No. of candidate values of the parameter $D_{LB}$ for Dual-LPUBoost.}\\
\end{tabular}
\end{center}
\end{table}

\begin{table*}[!ht]
    \centering
    \caption{Parameter settings for contending algorithms}\label{paramTab}
    \begin{threeparttable}
    \begin{tabular}{c|l|c} \hline
        Algorithm & \multicolumn{1}{c|}{Parameter settings} & Experiments \\
        \hline
        \multicolumn{3}{l}{\textbf{Baseline methods :}} \\ \hline
        AdaBoost \cite{freund1995desicion} & $T=10$ & Two-class \\ \hline
        AdaBoost.M2 \cite{freund1996experiments} & $T=10$ & Multi-class \\ \hline
        AdaMEC-Calib \cite{nikolaou2015calibrating} & $T=10$; $C_+ = \text{IR}$; $C_- = 1;$ & Two-class \\
        & Two random samplings of 80\% of the training data are used for training and calibration & Multi-class \\ \hline
        AdaBoost.NC \cite{wang2012multiclass} & $T=10$; $\lambda_{NC} \in \{2,9\}$ as per \cite{wang2012multiclass}; & Two-class \\
        & Random oversampling is done so as to equate the number of points in all classes & Multi-class \\ \hline
        \multicolumn{3}{l}{\textbf{LP based primal methods :}} \\ \hline
        LPAdaBoost \cite{grove1998boosting} & $T=10$ & Two-class, Multi-class \\ \hline
        LPUBoost \cite{leskovec2003linear} & $T=10$; $\nu \in \{0.1, 0.2\}$; $D = \frac{1}{\nu}$; $\beta \in \{2,4,8\}$ & Two-class \\ \hline
        \multicolumn{3}{l}{\textbf{LP based dual methods :}} \\ \hline
        Dual-LPAdaBoost \cite{grove1998boosting} & $T=10$; $\epsilon_{LPA} = 1\times10^{-16}$ & Two-class, Multi-class \\ \hline
        Dual-LPUBoost \cite{leskovec2003linear} & $T=10$; $\nu \in \{0.1, 0.2\}$; $D = \frac{1}{\nu}$; $\beta \in \{2,4,8\}$; $D_{LB} \in \{25, 50, 100\}$ & Two-class \\ \hline
        \multicolumn{3}{l}{\textbf{Proposed methods :}} \\ \hline
        LexiBoost & $T=10$ & Two-class, Multi-class \\ \hline
        Dual-LexiBoost & $T=10$ & Two-class, Multi-class \\ \hline
    \end{tabular}
    \begin{tablenotes}
    \item $^{1} C_+$ and $C_-$ denote the costs of false negatives and false positives, respectively.
    \item $^{2} \lambda_{NC}$ controls the strength of the penalty term in AdaBoost.NC.
    \item $^{3} \epsilon_{LPA}$ is the tolerance for primal-dual convergence in Dual-LPAdaBoost.
    \end{tablenotes}
    \end{threeparttable}
\end{table*}

\section{Experiments}\label{sec:exp}

In this section, we report the results of experiments conducted on two-class artificial datasets of varying specifications, two-class as well as multi-class real-world datasets, multi-class hyperspectral image classification, and multi-class classification of a class imbalanced subset of the ImageNet dataset. Our implementation of LexiBoost can be found at \url{https://github.com/Shounak-D/LexiBoost}.

\subsection{Competitors and Experimental Setup}\label{sec:contenders}

We compare our results with AdaBoost (the AdaBoost.M2 variant \cite{freund1996experiments} being used for multi-class datasets) which serves as a baseline, with AdaMEC-Calib which has recently been found to be quite effective for imbalanced datasets \cite{nikolaou2016cost}, and with the negative correlation based AdaBoost.NC in conjunction with random oversampling which has been shown to be effective on imbalanced data \cite{wang2012multiclass}. We also compare our results with those of the primal solutions (i.e. using classifiers already created by AdaBoost) as well as dual solutions to LPAdaBoost (because of its inherent ability to tackle imbalance in noise and outlier-free situations) and LPUBoost. The C4.5 decision tree \cite{quinlan2014c4} and the $k$-Nearest Neighbor ($k$NN) classifier are used as base learners. The experiments on the real-world datasets are reported with both C4.5 and $k$NN as base classifiers. Only the results with $k$NN as the base classifier are reported for the experiments on the artificial datasets, hyperspectral images and ImageNet, as the $k$NN based variants are observed to generally perform better in these cases \footnote{The corresponding results using C4.5 can be found in the supplementary document.}. The parameters for C4.5 are chosen as per \cite{quinlan2014c4} while the parameter $k$ for $k$NN is varied in the range $\{3, 5, 10\}$ for all sets of experiments except for hyperspectral image classification, where $k=10$ is used because of the relatively large size of the datasets. The parameter settings used for evaluating each of the contenders are summarized in Table \ref{paramTab}. The contending methods which are directly extendable to multi-class cases are used for the multi-class experiments, as indicated in Table \ref{paramTab}. 

The performance for the experiments is reported using the G-Mean \cite{kubat1997addressing}, AUC \cite{maloof2003auc}, and Avg-AUC \cite{hand2001simple} indexes. The G-Mean measure is calculated as the geometric mean of the individual class-wise accuracies. Since the G-Mean index has a high value only when the performance is good on all classes, it is useful for evaluating the performance for class imbalanced classification. The AUC index, on the other hand, measures the expected proportion of positive data samples which are more likely to be assigned to the positive class, compared to a randomly drawn negative sample. Since the AUC measure is only defined for two-class classification tasks, Hand and Till \cite{hand2001simple} proposed the Avg-AUC index as an extension of the AUC measure to multi-class problems using one-versus-one decomposition (i.e. average AUC over all possible pairings of the individual classes). Formal definitions of the indexes can be found in the supplementary document. The results are presented in the following sections using the average index values corresponding to the best parameter settings for each contending algorithm.

\subsection{Artificial Datasets}\label{sec:artif}

\begin{figure*}[!ht]
\vskip 0.2in
\begin{center}
\centerline{\includegraphics[width=\textwidth]{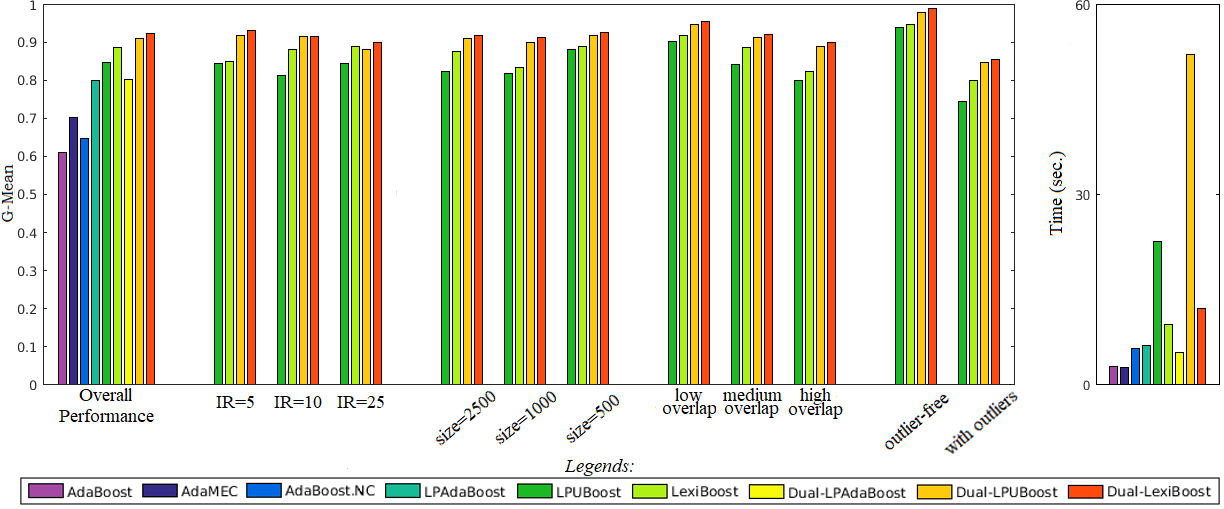}}
\caption{Average G-Mean values for $k$NN on artificial datasets: LexiBoost and Dual-LexiBoost are compared with the two best contenders.}\label{figkNN}
\end{center}
\vskip -0.2in
\end{figure*}

We create 27 two-class artificial datasets by sampling points from two distinct 5-dimensional standard normal distributions by varying the Imbalance Ratio (IR) in $\{5, 10, 25\}$ and the total size of the datasets in $\{500, 1000, 2500\}$. The overlap between the classes is also varied by keeping the centre for the minority class fixed at $[0, 0, 0, 0, 0]^{T}$ while the centre for the majority class is varied between $[3, 3, 3, 3, 3]^{T}$, $[1.7, 1.7, 1.7, 1.7, 1.7]^{T}$, and $[1.5, 1.5, 1.5, 1.5, 1.5]^{T}$. 27 analogous datasets with outliers are also created by replacing 10\% of the instances of each class with instances from the opposite class. The G-Mean values obtained for these 54 datasets using $k$NN as base classifier are summarized from different perspectives in Figure \ref{figkNN}. LPUBoost and Dual-LPUBoost, respectively being the best primal and dual methods apart from the proposed methods, are used for the comparison over varying IR, size, overlap and presence of outliers.

Figure \ref{figkNN} indicates that LexiBoost consistently performs better than the other primal techniques as well as the baselines AdaBoost, AdaMEC-Calib, and AdaBoost.NC. Similarly, Dual-LexiBoost consistently performs best among all the algorithms. In fact, Dual-LexiBoost even outperforms the exhaustive cost tuning methods, namely LPUBoost and Dual-LPUBoost. Moreover, the time required for training LexiBost and Dual-LexiBoost is much lower than that of their corresponding closest rivals LPUBoost and Dual-LPUBoost, and is comparable to that of the baseline techniques like AdaBoost. This points towards the effectiveness of the proposed framework for finding the best trade-off between classes, without cost tuning. An empirical validation of the ability of the LexiBoost framework to circumvent the requirement for cost tuning can be found in the supplementary material. It is important to note here that combining post-calibration with boundary-shifting enables AdaMEC-Calib to perform much better than AdaBoost, despite requiring the lowest training time among all the contenders. However, it does not perform as well as the other imbalance handling schemes, LPUBoost and Dual-LPUBoost. This is possibly because the final ensemble classifier learned by AdaBoost may be miscalibrated in a way which cannot be compensated for by shifting the decision threshold. Instead, carefully choosing the component weights may be able to help in such cases, as indicated by the relatively better performance of LPUBoost.

\begin{table*}[!th]
\caption{Summary of results on real-world datasets}\label{HResReal}
\begin{center}
\smallskip
\begin{tabular}{|c|c|c|l|ccccccccc|}
\cline{5-13}
\multicolumn{4}{c|}{} & \begin{sideways} AdaBoost \end{sideways} & \begin{sideways} AdaMEC-Calib \end{sideways} & \begin{sideways} AdaBoost.NC \end{sideways} & \begin{sideways} LPAdaBoost \end{sideways} & \begin{sideways} LPUBoost \end{sideways} & \begin{sideways} LexiBoost \end{sideways} & \begin{sideways} Dual-LPAdaBoost \;\; \end{sideways} & \begin{sideways} Dual-LPUBoost \end{sideways} & \begin{sideways} Dual-LexiBoost \end{sideways} \\
\hline
\multirow{14}{*}{\begin{sideways} Two-class datasets \end{sideways}} & \multicolumn{12}{|l|}{\textbf{C4.5 as base classifier :}} \\
\cline{2-13}
& \multirow{4}{*}{AUC} & \multicolumn{2}{l|}{Average Rank} & 6.80 & 4.90 & 8.07 & 5.30 & 4.27 & 2.70 & 7.23 & 3.17 & \textbf{2.57} \\
\cline{3-13}
& & \multicolumn{2}{l|}{Friedman Test} & \multicolumn{9}{l|}{$H_1$} \\
\cline{3-13}
& & \multirow{2}{*}{WSRT} & CN: LexiBoost & $H_1$ & $H_1$ & $H_1$ & $H_1$ & $H_1$ & - & $H_1$ & $H_0$ & $H_0$ \\
\cline{4-13}
& & & CN: Dual-LexiBoost & $H_1$ & $H_1$ & $H_1$ & $H_1$ & $H_0$ & $H_0$ & $H_1$ & $H_0$ & - \\
\cline{2-13}
& \multirow{4}{*}{G-Mean} & \multicolumn{2}{l|}{Average Rank} & 6.97 & 4.53 & 8.30 & 5.40 & 4.27 & 2.77 & 7.23 & 3.03 & \textbf{2.50} \\
\cline{3-13}
& & \multicolumn{2}{l|}{Friedman Test} & \multicolumn{9}{l|}{$H_1$} \\
\cline{3-13}
& & \multirow{2}{*}{WSRT} & CN: LexiBoost & $H_1$ & $H_1$ & $H_1$ & $H_1$ & $H_1$ & - & $H_1$ & $H_0$ & $H_0$ \\
\cline{4-13}
& & & CN: Dual-LexiBoost & $H_1$ & $H_1$ & $H_1$ & $H_1$ & $H_0$ & $H_0$ & $H_1$ & $H_0$ & - \\
\cline{2-13}
& \multicolumn{12}{|l|}{\textbf{$k$NN as base classifier :}} \\
\cline{2-13}
& \multirow{4}{*}{AUC} & \multicolumn{2}{l|}{Average Rank} & 7.80 & 6.00 & 6.97 & 6.10 & 4.73 & 3.10 & 5.27 & 2.63 & \textbf{2.40} \\
\cline{3-13}
& & \multicolumn{2}{l|}{Friedman Test} & \multicolumn{9}{l|}{$H_1$} \\
\cline{3-13}
& & \multirow{2}{*}{WSRT} & CN: LexiBoost & $H_1$ & $H_1$ & $H_1$ & $H_1$ & $H_1$ & - & $H_1$ & $H_0$ & $H_0$ \\
\cline{4-13}
& & & CN: Dual-LexiBoost & $H_1$ & $H_1$ & $H_1$ & $H_1$ & $H_1$ & $H_0$ & $H_1$ & $H_0$ & - \\
\cline{2-13}
& \multirow{4}{*}{G-Mean} & \multicolumn{2}{l|}{Average Rank} & 7.87 & 5.67 & 6.83 & 6.37 & 4.53 & 2.93 & 5.77 & 2.83 & \textbf{2.20} \\
\cline{3-13}
& & \multicolumn{2}{l|}{Friedman Test} & \multicolumn{9}{l|}{$H_1$} \\
\cline{3-13}
& & \multirow{2}{*}{WSRT} & CN: LexiBoost & $H_1$ & $H_1$ & $H_1$ & $H_1$ & $H_1$ & - & $H_1$ & $H_0$ & $H_0$ \\
\cline{4-13}
& & & CN: Dual-LexiBoost & $H_1$ & $H_1$ & $H_1$ & $H_1$ & $H_1$ & $H_0$ & $H_1$ & $H_1$ & - \\
\hline
\hline
\multirow{14}{*}{\begin{sideways} Multi-class datasets \end{sideways}} & \multicolumn{12}{|l|}{\textbf{C4.5 as base classifier :}} \\
\cline{2-13}
& \multirow{4}{*}{Avg-AUC} & \multicolumn{2}{l|}{Average Rank} & 3.70 & N/A & 3.80 & 5.10 & N/A & \textbf{1.80} & 4.70 & N/A & 1.90 \\
\cline{3-13}
& & \multicolumn{2}{l|}{Friedman Test} & \multicolumn{9}{l|}{$H_1$} \\
\cline{3-13}
& & \multirow{2}{*}{WSRT} & CN: LexiBoost & $H_1$ & N/A & $H_1$ & $H_1$ & N/A & - & $H_1$ & N/A & $H_0$ \\
\cline{4-13}
& & & CN: Dual-LexiBoost & $H_1$ & N/A & $H_1$ & $H_1$ & N/A & $H_0$ & $H_1$ & N/A & - \\
\cline{2-13}
& \multirow{4}{*}{G-Mean} & \multicolumn{2}{l|}{Average Rank} & 3.75 & N/A & 3.80 & 4.80 & N/A & 2.55 & 4.00 & N/A & \textbf{2.10} \\
\cline{3-13}
& & \multicolumn{2}{l|}{Friedman Test} & \multicolumn{9}{l|}{$H_1$} \\
\cline{3-13}
& & \multirow{2}{*}{WSRT} & CN: LexiBoost & $H_0$ & N/A & $H_0$ & $H_1$ & N/A & - & $H_1$ & N/A & $H_0$ \\
\cline{4-13}
& & & CN: Dual-LexiBoost & $H_1$ & N/A & $H_1$ & $H_1$ & N/A & $H_0$ & $H_1$ & N/A & - \\
\cline{2-13}
& \multicolumn{12}{|l|}{\textbf{$k$NN as base classifier :}} \\
\cline{2-13}
& \multirow{4}{*}{Avg-AUC} & \multicolumn{2}{l|}{Average Rank} & 4.40 & N/A & 3.95 & 4.60 & N/A & 2.50 & 4.30 & N/A & \textbf{1.25} \\
\cline{3-13}
& & \multicolumn{2}{l|}{Friedman Test} & \multicolumn{9}{l|}{$H_1$} \\
\cline{3-13}
& & \multirow{2}{*}{WSRT} & CN: LexiBoost & $H_1$ & N/A & $H_0$ & $H_1$ & N/A & - & $H_1$ & N/A & $H_1$ \\
\cline{4-13}
& & & CN: Dual-LexiBoost & $H_1$ & N/A & $H_1$ & $H_1$ & N/A & $H_1$ & $H_1$ & N/A & - \\
\cline{2-13}
& \multirow{4}{*}{G-Mean} & \multicolumn{2}{l|}{Average Rank} & 4.10 & N/A & 4.25 & 4.75 & N/A & 2.70 & 3.60 & N/A & \textbf{1.60} \\
\cline{3-13}
& & \multicolumn{2}{l|}{Friedman Test} & \multicolumn{9}{l|}{$H_1$} \\
\cline{3-13}
& & \multirow{2}{*}{WSRT} & CN: LexiBoost & $H_1$ & N/A & $H_1$ & $H_1$ & N/A & - & $H_0$ & N/A & $H_1$ \\
\cline{4-13}
& & & CN: Dual-LexiBoost & $H_1$ & N/A & $H_1$ & $H_1$ & N/A & $H_1$ & $H_1$ & N/A & - \\
\hline
\multicolumn{6}{l}{$H_0:$ Contenders perform similarly} & \multicolumn{7}{r}{WSRT $:$ Wilcoxon Signed Rank Test} \\
\multicolumn{6}{l}{$H_1:$ Significant difference among contenders} & \multicolumn{7}{r}{CN $:$ Control method} \\
\multicolumn{6}{l}{Best values shown in \textbf{boldface}} & \multicolumn{7}{r}{N/A $:$ Not used for multi-class experiments} \\
\end{tabular}
\end{center}
\end{table*}

\subsection{Real-world Datasets}\label{sec:real}

For the experiments on real-world imbalanced datasets, we use 15 two-class and 10 multi-class datasets with varying degrees of imbalance from the KEEL repository \cite{triguero2017} (see supplementary document for details). All the methods are compared on the two-class datasets while only AdaBoost (the AdaBoost.M2 variant), LPAdaBoost, LexiBoost, Dual-LPAdaBoost and Dual-LexiBoost are compared on the multi-class datasets (as the other methods are not directly adaptable to multi-class problems).

\subsubsection{Two-class Classification}\label{sec:twoClassRes}

The performance over the 15 two-class real-world imbalanced datasets is summarized in Table \ref{HResReal} for C4.5 and $k$NN as the base classifier, respectively. The results are summarized in terms of average ranks, Friedman test hypotheses, and Wilcoxon signed rank test hypotheses \cite{wilcoxon1945individual}. The Friedman test \cite{friedman1937use} is used to ascertain whether there is significant difference among the performances of the various contenders. Since the Friedman test suggests significant difference in all cases, we further employ the signed rank test to investigate the pair-wise differences between the contenders. It is seen that Dual-LexiBoost achieves the best rank, followed by LexiBoost and Dual-LPUBoost. The performances of Dual-LexiBoost, LexiBoost and Dual-LPUBoost are found to be statistically equivalent in terms of both G-Mean and AUC while all the other contenders are found to perform significantly worse than both Dual-LexiBoost and LexiBoost for both choices of base classifiers. This indicates that the proposed framework has the capacity to perform at least as good as exhaustive cost tuning methods, even on real-world datasets. The lower average rankings of Dual-LexiBoost (relative to LexiBoost) indicate that the dual formulation for adapting instance weights is indeed useful for achieving better performance.

\subsubsection{Multi-class Classification}\label{sec:multiClassRes}

The performance of AdaBoost.M2, AdaBoost.NC, LPAdaBoost, LexiBoost, Dual-LPAdaBoost and Dual-LexiBoost (on the multi-class real-world datasets having varying number of classes and degree of imbalance) is summarized for C4.5 and $k$NN respectively in the lower half of Table \ref{HResReal}. The Friedman test detects significant difference among the contenders in all cases. Hence, the Wilcoxon signed rank test, along with average ranks are used for summarizing the performance. The overall performance is the best for Dual-LexiBoost followed by LexiBoost. The performance of Dual-LexiBoost is significantly better than that of all other methods in terms of both indexes. This indicates that the dual method of instance weight adaptation is useful for generating better balance between the classes for multi-class datasets as well.

\begin{figure*}[!t]
\centering
\subfloat[Part of KSC image showing minority classes.]{\includegraphics[width=1.35in]{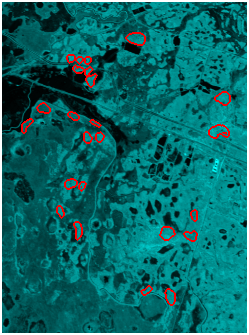}
\label{figKSCchannel}}
\subfloat[Ground truth.]{\includegraphics[width=1.332in]{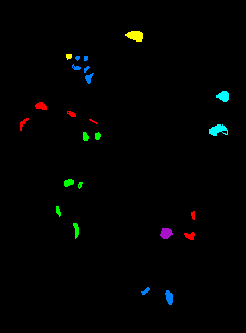}
\label{figKSCgt}}
\subfloat[LexiBoost.]{\includegraphics[width=1.347in]{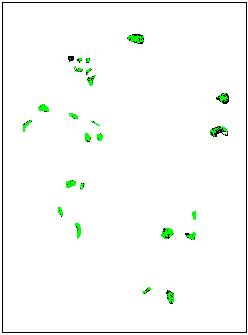}
\label{figKSCfin}}
\subfloat[Dual-LexiBoost.]{\includegraphics[width=1.35in]{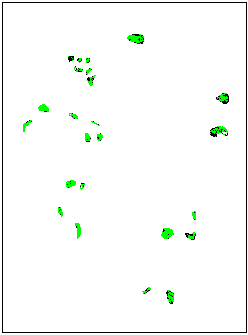}
\label{figKSCdfin}}
\hfill
\subfloat[AdaBoost.M2.]{\includegraphics[width=1.35in]{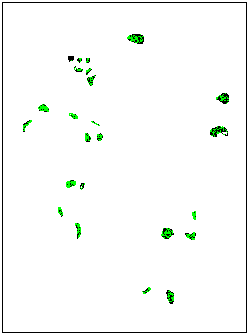}
\label{figKSCada}}
\subfloat[AdaBoost.NC.]{\includegraphics[width=1.35in]{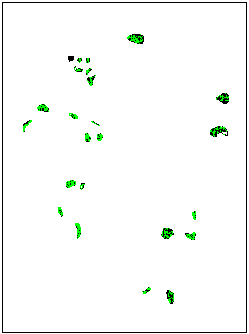}
\label{figKSCnc}}
\subfloat[LPAdaBoost.]{\includegraphics[width=1.35in]{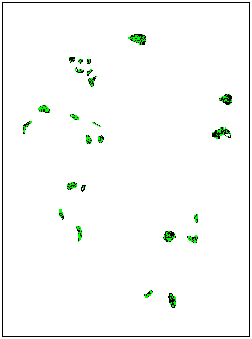}
\label{figKSClpa}}
\subfloat[Dual-LPAdaBoost.]{\includegraphics[width=1.35in]{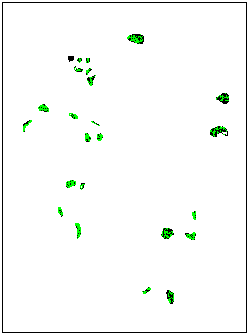}
\label{figKSCdlpa}}
\caption{Comparison of the contending methods showing better performance of the proposal on minority classes. Misclassified points are shown in black and correctly-classified points are shown in green.}
\label{figVarious}
\end{figure*}

\begin{table}[!th]
\caption{Summary of results on hyperspectral images and ImageNet}\label{HResImg}
\begin{center}
\smallskip
\begin{tabular}{|c|c|c|cccccc|}
\cline{4-9}
\multicolumn{3}{c|}{} & \begin{sideways} AdaBoost \end{sideways} & \begin{sideways} AdaBoost.NC \end{sideways} & \begin{sideways} LPAdaBoost \end{sideways} & \begin{sideways} LexiBoost \end{sideways} & \begin{sideways} Dual-LPAdaBoost \;\; \end{sideways} & \begin{sideways} Dual-LexiBoost \end{sideways} \\
\hline
\multicolumn{9}{|l|}{\textbf{Hyperspectral Images :}} \\
\hline
\multirow{7}{*}{\begin{sideways} Avg-AUC \end{sideways}} & \multicolumn{2}{c|}{Avg. Rank} & 3.40 & 4.60 & 5.50 & 2.40 & 4.10 & \textbf{1.00} \\
\cline{2-9}
& \multirow{3}{*}{\shortstack{CN: Lexi-\\Boost}} & W & 0 & 0 & 0 & - & 0 & 1 \\
& & T & 2 & 1 & 1 & - & 2 & 1 \\
& & L & 0 & 1 & 1 & - & 0 & 0 \\
\cline{2-9}
& \multirow{3}{*}{\shortstack{CN: Dual-\\LexiBoost}} & W & 0 & 0 & 0 & 0 & 0 & - \\
& & T & 1 & 0 & 0 & 1 & 1 & - \\
& & L & 1 & 2 & 2 & 1 & 1 & - \\
\hline
\multirow{7}{*}{\begin{sideways} G-Mean \end{sideways}} & \multicolumn{2}{c|}{Avg. Rank} & 3.40 & 4.70 & 5.50 & 2.40 & 4.00 & \textbf{1.00} \\
\cline{2-9}
& \multirow{3}{*}{\shortstack{CN: Lexi-\\Boost}} & W & 0 & 0 & 0 & - & 0 & 1 \\
& & T & 2 & 0 & 1 & - & 1 & 1 \\
& & L & 0 & 2 & 1 & - & 1 & 0 \\
\cline{2-9}
& \multirow{3}{*}{\shortstack{CN: Dual-\\LexiBoost}} & W & 0 & 0 & 0 & 0 & 0 & - \\
& & T & 1 & 0 & 0 & 1 & 1 & - \\
& & L & 1 & 2 & 2 & 1 & 1 & - \\
\hline
\hline
\multicolumn{9}{|l|}{\textbf{ImageNet :}} \\
\hline
\multirow{7}{*}{\begin{sideways} Avg-AUC \end{sideways}} & \multicolumn{2}{c|}{Avg. Rank} & 3.67 & 3.00 & 5.67 & 2.67 & 5.00 & \textbf{1.00} \\
\cline{2-9}
& \multirow{3}{*}{\shortstack{CN: Lexi-\\Boost}} & W & 0 & 0 & 0 & - & 0 & 1 \\
& & T & 3 & 3 & 2 & - & 2 & 2 \\
& & L & 0 & 0 & 1 & - & 1 & 0 \\
\cline{2-9}
& \multirow{3}{*}{\shortstack{CN: Dual-\\LexiBoost}} & W & 0 & 0 & 0 & 0 & 0 & - \\
& & T & 0 & 1 & 0 & 1 & 0 & - \\
& & L & 3 & 2 & 3 & 2 & 3 & - \\
\hline
\multirow{7}{*}{\begin{sideways} G-Mean \end{sideways}} & \multicolumn{2}{c|}{Avg. Rank} & 4.33 & 3.00 & 5.00 & 3.00 & 4.67 & \textbf{1.00} \\
\cline{2-9}
& \multirow{3}{*}{\shortstack{CN: Lexi-\\Boost}} & W & 0 & 0 & 0 & - & 0 & 3 \\
& & T & 2 & 3 & 2 & - & 1 & 0 \\
& & L & 1 & 0 & 1 & - & 2 & 0 \\
\cline{2-9}
& \multirow{3}{*}{\shortstack{CN: Dual-\\LexiBoost}} & W & 0 & 0 & 0 & 0 & 0 & - \\
& & T & 0 & 0 & 0 & 0 & 0 & - \\
& & L & 3 & 3 & 3 & 3 & 3 & - \\
\hline
\multicolumn{9}{l}{W, T, L$:$ Wilcoxon rank-sum test Win, Tie, Loss counts, resp.} \\
\multicolumn{5}{l}{Best values shown in \textbf{boldface}} & \multicolumn{4}{r}{CN $:$ Control method} \\
\end{tabular}
\end{center}
\end{table}

\subsection{Hyperspectral Image Classification}\label{sec:hyper}

Hyperspectral image classification has been listed, in a recent survey by Krawczyk \cite{Krawczyk2016}, as one of the key practical application areas where multi-class imbalance naturally arises. Therefore, in this section, we test the effectiveness of the proposed techniques for this application. For the experiments on hyperspectral image classification, we use the Samson and Jasper ridge images from \cite{zhu2014hyper}, the Kennedy Space Center (KSC) image from \cite{bandos2009hyper}, and the Salinas A and Indian pines scenes from \cite{hyperWebsite}. For the images which contain unlabeled pixels, the training and testing is only undertaken on the labeled pixels as per \cite{sun2015hyper}. We only use $k$NN (with $k=10$) as the base classifier for these experiments because of the large size of these datasets. The results are summarized in Table \ref{HResImg} in terms of the average ranking and the rank-sum test \cite{wilcoxon1945individual, mann1947test} win, tie, loss counts for both Avg-AUC and G-Mean. The Kruskal-Wallis test \cite{kruskal1952use} is used to detect difference among all contenders for each dataset. The wins, ties, and losses (of the contender against the control, viz. LexiBoost or Dual-LexiBoost) are only counted on the datasets which have significant differences among the contenders according to the Kruskal-Wallis test. We use the Kruskal-Wallis and rank-sum tests for each dataset, instead of using the Frideman and signed rank tests across datasets, owing to the limited number of datasets. The effectiveness of the proposed methods is attested to by their low average ranks and is also visible from the illustration in Figure \ref{figVarious} for the KSC image.

\begin{figure}[h]
\begin{center}
\centerline{\includegraphics[width=\columnwidth]{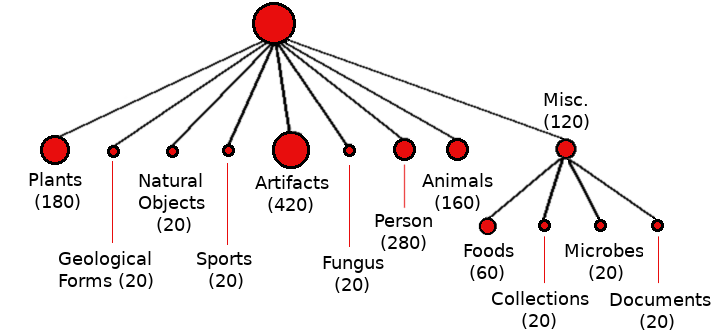}}
\caption{ImageNet subset hierarchy.}\label{fig:imageNetTree}
\end{center}
\vspace{-7mm}
\end{figure}

\subsection{ImageNet Classification}\label{sec:imagenet} 

One of the more challenging applications of pattern recognition is the classification of natural images. Uncurated natural image datasets are inherently class imbalanced. Moreover, the class distributions for such datasets are generally complex, making learning algorithms more sensitive to issues such as cost set tuning and outlier regularization. Therefore, in this section, we undertake the classification of imbalanced subsets of the popular ImageNet dataset \cite{imageNet}. We prepare 3 datasets, namely ImageNet8, ImageNet9, and ImageNet12 for this purpose. The datasets are prepared by randomly choosing images corresponding to the 8 principal subtrees of the ImageNet dataset as well as the miscellaneous subtrees, viz. Foods, Collections, Documents, and Microorganisms, from the ImageNet 2011 Fall Release. The number of images collected from each of the subtrees corresponds to about 2\% of the number of synsets contained within the subtree in question, with the constraint that at least 20 images must be chosen from each subtree. The data sampling hierarchy thus obtained is illustrated in Figure \ref{fig:imageNetTree}. The dataset ImageNet8 is prepared by only combining the samples from the 8 principal subtrees and not including the miscellaneous images, giving rise to a dataset containing 1120 images. The ImageNet9 dataset adds to the complexity of the classification task by appending the 120 miscellaneous images as a single class, resulting in a dataset of size 1240. The complexity is increased further in the ImageNet12 dataset as the images belonging to the miscellaneous subtrees Foods, Collections, Documents, and Microorganisms are classified into 4 different classes corresponding to these subtrees. In keeping with the state-of-the-art in feature representation of images, we derive a 2048-dimensional deep feature space representation of each image from the final global average pooling layer of the Inception-v3 deep neural network \cite{szegedy2016rethinking}. The results achieved by AdaBoost.M2, AdaBoost.NC, LPAdaBoost, LexiBoost, Dual-LPAdaBoost and Dual-LexiBoost are also summarized in Table \ref{HResImg}. Yet again, Dual-LexiBoost is observed to achieve the best rank followed by LexiBoost. The fact that Dual-LexiBoost (unlike LexiBoost) exhibits tie counts of zero against all other contenders in terms of G-Mean indicates that the dual formulation can generate proper instance weights to improve the performance on all classes (as opposed to only some of the classes by LexiBoost) for multi-class datasets.

\section{Conclusions}\label{sec:concl}

Based on the understanding that the choice of component classifier weights for boosting can be thought of as a game of \emph{Tug of War} between the classes in the margin space, we introduce the reader to a two-staged LxLP framework for handling class imbalance. The proposed framework, called LexiBoost, introduces a novel regularization scheme which offers an advantages over the traditional slack-variable-reliant scheme, due to the fact that the proposed scheme does not require to undertake expensive cost set tuning which has been the norm for imbalanced classification till date. Hence, the proposed framework also facilitates easy extension to multi-class problems. This makes LexiBoost directly applicable to both two-class as well as multi-class tasks. We also derive the dual algorithm corresponding to the proposed method, called Dual-LexiBoost. Experiments conducted on artificial datasets, real-world imbalanced datasets and hyperspectral images suggest that the proposed methods exhibit greater immunity to class imbalance, overlap, size of the dataset, as well as the presence of outliers. Dual-LexiBoost, owing to its ability to generate suitable point-specific weights, generally performs better than the primal method. In the near future, the authors plan to extend the proposed framework to single-class classification along the lines of \cite{ratsch2000svm}.

\section*{Acknowledgement}
We would like to thank Anubhav Agrawal, final year student pursuing B.Tech. in Electronics and Electrical Engineering from the Indian Institute of Technology, Guwahati, India, for helping with the computer implementation of some of the methods used in our experiments.

\bibliographystyle{IEEEtran}
\bibliography{refsTkde}

%

\vspace{-1cm}
\begin{IEEEbiography}[{\includegraphics[width=1in,height=1.25in,clip,keepaspectratio]{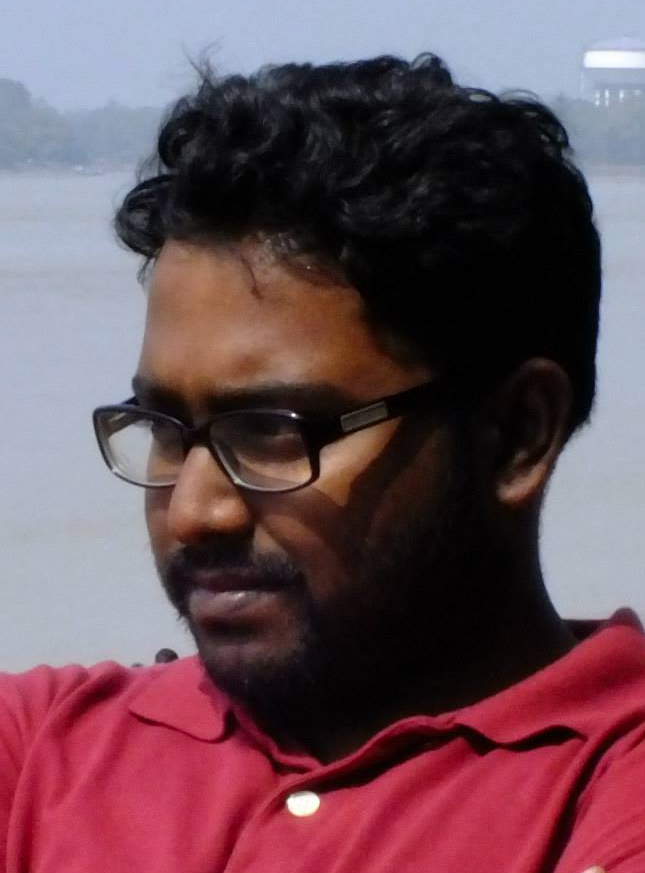}}]{Shounak Datta}
received his B.Tech. degree in Electronics and Communication Engineering from West Bengal University of Technology, Kolkata, India in 2011, and M.E. in Electronics and Telecommunication Engineering from Jadavpur University, Kolkata, India in 2013. He is currently pursuing a Ph.D. in Computer Science from Indian Statistical Institute, Kolkata, India. His research interests include imbalanced classification, learning with missing features, multi-objective optimization in machine learning, etc.
\end{IEEEbiography}

\vspace{-1.2cm}
\begin{IEEEbiography}[{\includegraphics[width=1in,height=1.1in,clip,keepaspectratio]{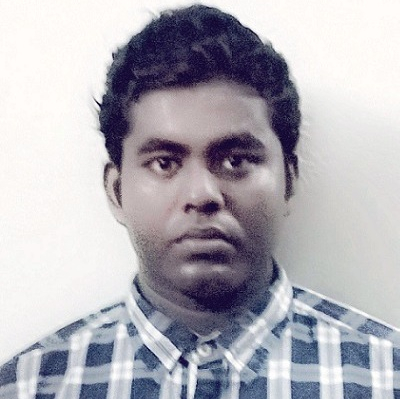}}]{Sayak Nag} has recently completed his B.E. in Instrumentation and Electronics Engineering from the Jadavpur University, Kolkata, India. His research interests include ensembles classifiers, support vector machines, neural networks, multi-objective optimization, and machine learning in general.
\end{IEEEbiography}

\vspace{-1.5cm}
\begin{IEEEbiography}[{\includegraphics[width=1in,height=1.25in,clip,keepaspectratio]{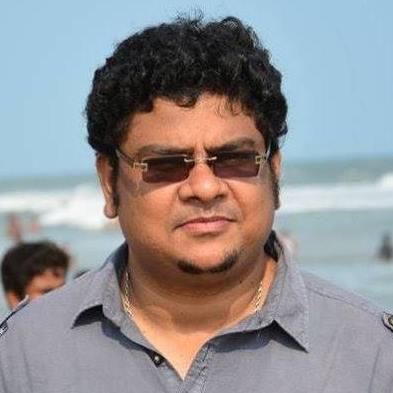}}]{Swagatam Das}
is currently serving as an associate professor at the Electronics and Communication Sciences Unit, Indian Statistical Institute, Kolkata, India. He has published more than 250 research articles in peer-reviewed journals and international conferences. He is the founding co-editor-in-chief of “Swarm and Evolutionary Computation”, an international journal from Elsevier. Dr. Das has 15,000+ Google Scholar citations and an H-index of 60 till date.
\end{IEEEbiography}

\end{document}